\documentclass[10pt, a4paper]{article}
\usepackage{lrec}
\usepackage{multibib}
\newcites{languageresource}{Language Resources}
\usepackage{graphicx}
\usepackage{tabularx}
\usepackage{soul}

\usepackage{epstopdf}
\usepackage[latin1]{inputenc}

\usepackage{hyperref}
\usepackage{xstring}

\title{Page Stream Segmentation with Convolutional Neural Nets Combining Textual and Visual Features}

\name{Gregor Wiedemann, Gerhard Heyer}

\address{Department of Computer Science \\
         Leipzig University, Germany \\
         gregor.wiedemann@uni-leipzig.de, heyer@informatik.uni-leipzig.de \\}

\abstract{
In recent years, (retro-)digitizing paper-based files became a major undertaking for private and public archives as well as an important task in electronic mailroom applications. 
As a first step, the workflow involves scanning and Optical Character Recognition (OCR) of documents. 
Preservation of document contexts of single page scans is a major requirement in this context. 
To facilitate workflows involving very large amounts of paper scans, page stream segmentation (PSS) is the task to automatically separate a stream of scanned images into multi-page documents.
In a digitization project together with a German federal archive, we developed a novel approach based on convolutional neural networks (CNN) combining image and text features to achieve optimal document separation results. Evaluation shows that our PSS architecture achieves an accuracy up to 93~\% which can be regarded as a new state-of-the-art for this task.
\\ \newline  }

\begin{document}

\maketitleabstract

\section{Introduction}

For digitization of incoming mails in business contexts as well as for retro-digitizing archives, batch scanning of documents can be a major simplification of the processing workflow. In this scenario, scanned images of multi-page documents arrive at a document management system as an ordered stream of single pages lacking information on document boundaries. 
Page stream segmentation (PSS) then is the task of dividing the continuous document stream into sequences of pages that represent single physical documents.\footnote{The task is also referred to as Document Flow Segmentation or Document Separation.}

Applying a fully automated approach of document page segmentation can be favorable over manually separating and scanning documents, especially in contexts of very large data sets which need to be separated \cite{Gallo.2016}.
In a joint research project together with a German research archive, we supported the task of retro-digitization of a paper archive consisting of circa one million pages put on file between 1922 and 2010 \cite{Isemann.2014}. 
The collection contains documents of varying content, types and lengths around the topic of ultimate disposal of nuclear waste, mostly administrative letter correspondence and research reports, but also stock lists, meeting minutes and email printouts. 
The 1M pages were archived in roughly 20.000 binders which were batch-scanned due to limited manual capacities for separating individual documents. The long time range of archived material affects document quality, proliferation of layout standards, different fonts and the use of hand-written texts. All these circumstances pose severe challenges to OCR as well as to page stream segmentation (PSS).

In this article, we introduce our approach to PSS comparing (linear) support vector machines (SVM) and convolutional neural networks (CNN). For the first time for this task, we combine textual and visual features into one network to achieve most-accurate results. The upcoming section \ref{sec:related_work} elaborates on related work. In section \ref{sec:dataset} we describe our dataset together with one reference dataset for this task. In section \ref{sec:approach} we introduce our neural network based architecture for PSS. 
As a baseline, we introduce an SVM-based model solely operating on text features. Then, we introduce CNN for PSS on text and image data separately as well as in a combined architecture. Section \ref{sec:eval} presents a quantitative and a qualitative evaluation of the approach on the two datasets.

\section{Related work}
\label{sec:related_work}

Page stream segmentation is related to a series of other tasks concerned with digital document management workflows. Table~\ref{tab:related_work} summarizes important characteristics of recent works in this field. A common task related to PSS is document image classification (DIC) in which typically visual features (pixels) are utilized to classify scanned document representations into categories such as ``invoice", ``letter", ``certificate" etc. Category systems can become quite large and complex. \cite{Gordo.2013} summarize different approaches in a survey article on PSS and DIC. 

In \cite{Gallo.2016}, PSS is performed on top of the results from a DIC process. Pages from the stream are segmented each time the DIC system detects a change of class labels between consecutive page images. 
This approach can only be successful in case there are alternating types of documents in the sequential stream. Often, this cannot be guaranteed, especially in case of small document category systems. 

Alternative approaches seek to identify document boundaries explicitly. Such approaches are proposed in \cite{Daher.2014} and \cite{Agin.2015} where each individual image of the sequence is classified as either continuity of the same document (SD) or beginning of a new document (ND). For this binary classification, \cite{Daher.2014} rely on textual features extracted from OCR-results and classify pages with SVM and multi-layer perceptrons (MLP). \cite{Agin.2015} employ bag of visual words (BoVW) and font information obtained from OCR as features, and test performance with three binary classifiers (SVM, Random Forest, and MLP).

The recent state-of-the-art for DIC is achieved by \cite{Gallo.2016}, \cite{Harley.2015} and \cite{Noce.2016} who employ Deep Learning with Convolutional Neural Networks to identify document classes. While the former two employ only visual features, the latter study uses both, visual and text features for DIC. For this, class-specific key terms are extracted from the OCR-ed training documents and highlighted with correspondingly colored boxes in the document images. Then, a CNN is applied to learn document classes from these images augmented with textual information highlighting.

Although with \cite{Gallo.2016} there is already one study employing neural network technology not only for DIC but also for PSS, their approach was not applicable to our project for two reasons. First, as mentioned earlier, they perform PSS only indirectly based on changing class labels of consecutive pages. Since we only have 17 document categories and a majority of them belong to one category ("letter"), we need to perform direct separation of the page stream by classifying each page into either SD or ND. Second, quality and layout of our data is extremely heterogeneous due to the long time period of document creation. We expect a lowered performance by solely relying on visual features for separation. Therefore, taking the previous work of \cite{Gallo.2016} as a starting point, we propose our approach for direct PSS as a binary classification task combining textual features and visual features using deep neural networks. We compare this architecture against a baseline comprising an SVM classifier solely relying on textual features.

\begin{table*}
\centering
\caption{Recent work on page stream segmentation}
\label{tab:related_work}
\begin{tabular}{lccccll}
\hline
\textbf{Authors}     & \textbf{PSS} & \textbf{DIC} & \textbf{Visual Features} & \textbf{Text Features} & \textbf{Architecture} & \textbf{Accuracy} \\ \hline
Daher; Belaid (2014) & X            &              &                          & X                      & SVM, MLP              & F = 0.8 - 0.9     \\
Agin et al. (2015)   & X            &              & X                        & (X, fonts)                    & SVM, RF, MLP          & F = 0.89          \\
Harley et al. (2015) &              & X            & X                        &                        & CNN                   & A = 0.76 - 0.90   \\
Noce et al. (2016)   &              & X            & X                        & X                      & CNN                   & A = 0.8 - 0.9     \\
Gallo et al. (2016)  & (X, indirect) & X            & X                        &                        & CNN+MLP               & A = 0.88          \\ \hline
Our approach         & X            &              & X                        & X                      & SVM                   &                   \\
                     &              &              &                          &                        & CNN+MLP               & see Section \ref{sec:eval}                  \\ \hline
\end{tabular}
\end{table*}

\section{Datasets}
\label{sec:dataset}


We evaluate our approach on two datasets, one sample from the German archive data of our project context, and one public resource of annotated document scans from U.S. tobacco companies. 

\subsection{German archive data}

The German dataset consists of a variety of document classes from a very long time frame. Most of the documents were archived between the mid-1960s and 2010. Due to this, OCR-quality, document lengths, layout standards as well as used fonts differ widely. 

\begin{figure}
  \centering
\fbox{\includegraphics[width=4.9em, height=6.9em]{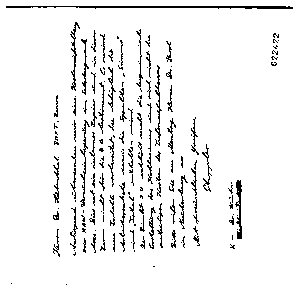}}\vspace{0.2em}
\fbox{\includegraphics[width=4.9em, height=6.9em]{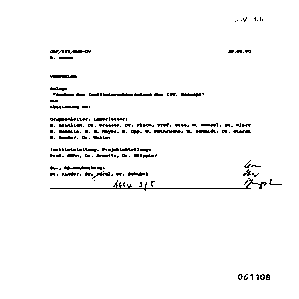}}
\fbox{\includegraphics[width=4.9em, height=6.9em]{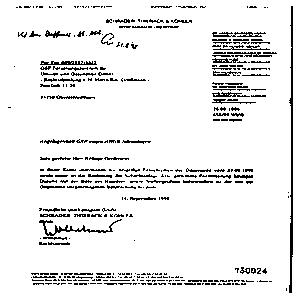}}
\fbox{\includegraphics[width=4.9em, height=6.9em]{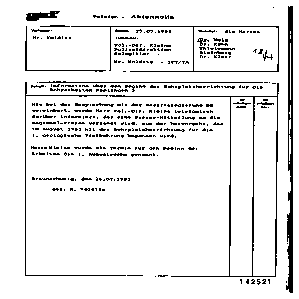}}
\fbox{\includegraphics[width=4.9em, height=6.9em]{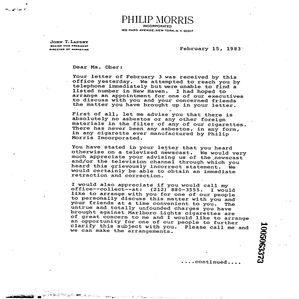}}
\fbox{\includegraphics[width=4.9em, height=6.9em]{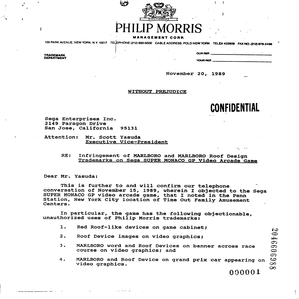}}
\fbox{\includegraphics[width=4.9em, height=6.9em]{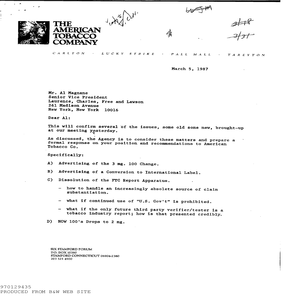}}
\fbox{\includegraphics[width=4.9em, height=6.9em]{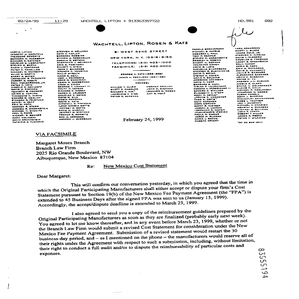}}
  \caption{Examples for first pages (class \textit{new document}); from Archive22k (above) and Tobacco800 (below).}
  \label{fig:scaled_pics1}
\end{figure}

\begin{figure}
  \centering
\fbox{\includegraphics[width=4.9em, height=6.9em]{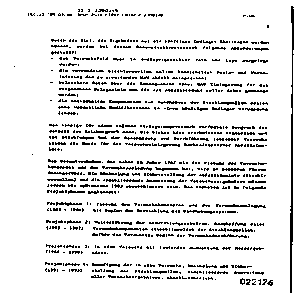}} \vspace{0.2em}
\fbox{\includegraphics[width=4.9em, height=6.9em]{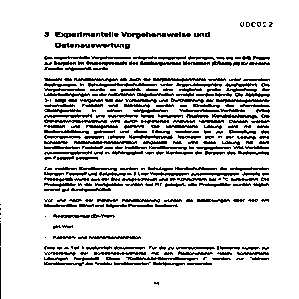}}
\fbox{\includegraphics[width=4.9em, height=6.9em]{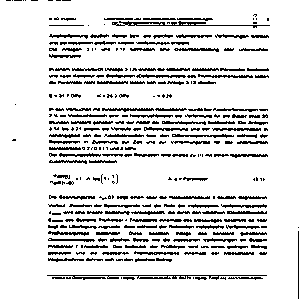}}
\fbox{\includegraphics[width=4.9em, height=6.9em]{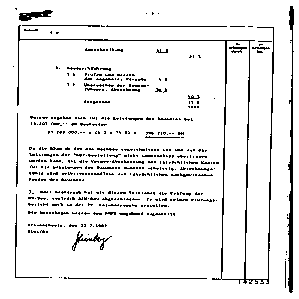}}
\fbox{\includegraphics[width=4.9em, height=6.9em]{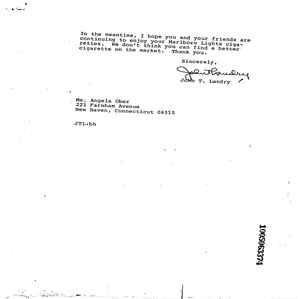}}
\fbox{\includegraphics[width=4.9em, height=6.9em]{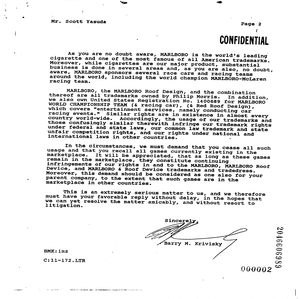}}
\fbox{\includegraphics[width=4.9em, height=6.9em]{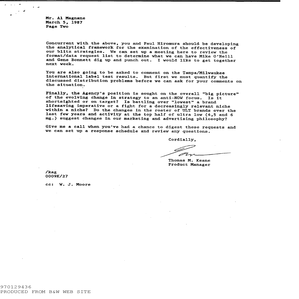}}
\fbox{\includegraphics[width=4.9em, height=6.9em]{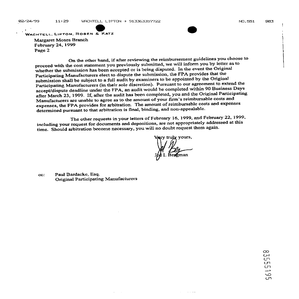}}
\caption{Examples for subsequent pages (class \textit{same document}); from Archive22k (above) and Tobacco800 (below).}
  \label{fig:scaled_pics2}
\end{figure}

After batch scanning, about 40~\% of all binders from the German research archive have been manually separated into documents and annotated with document categories. The manually separated documents can serve as a ground truth for our experiments on model selection and feature engineering for automatic page stream segmentation.
For these experiments, we randomly selected 100 binders from the set of all manually separated binders. The binders represent 100 ordered streams of scanned pages, in total consisting of  22,741 pages. 80 of the selected binders containing 17,376 pages were taken as a training set, 20 binders with 5095 pages were taken as test set. 
Scanned pages were resampled to a size of 224 $\times$ 224 pixels and color-converted to black and white with the OTSU binarization method \cite{Otsu.1979}. 
The upper lines in Fig. \ref{fig:scaled_pics1} and \ref{fig:scaled_pics2} show examples of first pages, resp. subsequent pages from documents.

From original document scans, text information was extracted by optical character recognition (OCR). In the following, this dataset is referred to as \emph{Archive22k}.

\subsection{Tobacco800}

As a second evaluation set, we run our classification process on the \emph{Tobacco800} document image database \cite{Lewis.2006}. The dataset allows comparing the performance of our approach to other recent studies. 

The Tobacco800 dataset is a small annotated subset of the Truth Tobacco Industry Documents, a collection of more than 14 million documents originating from seven major U.S. tobacco industry organizations  dealing with their research, manufacturing, and marketing during the last decades. The documents had to be publicly released due to lawsuits in the United States.

The annotated subset for our experiments is composed of 1,290 document images sampled from the original corpus. Similar to the German dataset, it contains multi-page documents of different types (e.g. letters, invoices, hand-written documents) and thus is well suited for evaluation of our task. Samples from the Tobacco dataset were also used in \cite{Harley.2015} and \cite{Noce.2016}. Again, we extract text information for each page via OCR from the original page scans, OTSU-binarize them to a black/white color palette and resize them to a 224 $\times$ 224 pixel resolution.
The lower lines in Fig. \ref{fig:scaled_pics1} and \ref{fig:scaled_pics2} show examples of first pages, resp. subsequent pages from Tobacco800 documents.

\vspace{1em}
As the example pages show, both collections share similarities in their visual appearance. First pages compared to subsequent ones may contain distinct header elements. But in general, the human observer has difficulties to identify clear layout patterns discriminating between both classes, especially for the Archive22k documents. Therefore, visual features alone may not be sufficient for accurate PSS.

Regarding their textual content, the two datasets share certain similarities but also differ with respect to language, size, and creatorship. 
Both have in common that they cover long time periods and are thematically located within a rather narrow domain (nuclear waste disposal, tobacco industry). 
Nonetheless, they largely differ 
with respect to characteristics of content creators. On the one hand, there is a state-run research library archiving material from a wide variety of actors, while on the other hand there are internal documents from a rather small set of business actors with corporate design standards. Due to this, we expect different performance from textual and visual features for PSS on both datasets.

\section{Binary classification for PSS}
\label{sec:approach}

Analogue to \cite{Daher.2014} and \cite{Agin.2015}, we approach PSS as a binary classification task on single pages from a data stream. Pages are classified into either continuity of the \emph{same document} (SD) or beginning of a \emph{new document} (ND). For classification, we compare two architectures: SVM with specifically engineered text features (\ref{subsec:svm}) and a combination of CNN and MLP with both, textual and visual features (\ref{subsec:cnn}).

\subsection{Baseline: SVM on text features}
\label{subsec:svm}

As a baseline, we use linear text classification together with specifically engineered features for PSS. For this first step, we rely on SVM with a linear kernel\footnote{We use the Liblinear library by \cite{Fan.2008}}. This learning algorithm has proven to be very efficient for binary classification problems with sparse and large feature spaces \cite{Joachims.1998} and is computationally much faster than neural network architectures.\footnote{We refrain from using image features in this architecture, because pixel features are not supposed to be linearly separable. First experiments confirmed that pixel features do not contribute discriminative information on top of text features to the linear SVM for our task. Of course, we could use a different SVM kernel for image classification. But, very likely we would lose the advantage of computational speed. Due to this, we stick to text features for our baseline method.}

We extract four types of features from the OCR-ed text data of the single pages. 

\paragraph{Unigrams:} Page texts were tokenized and resulting tokens reduced to their word stem. We further replaced digits in tokens with a \#-character and pruned types from the vocabulary which occurred less than 3 times (Tobacco800), resp. 10 times (Archive22k). Pruning was applied to maintain manageable vocabulary sizes and reduce noise from infrequent events in the data. Different thresholds for feature pruning were chosen with respect to different collection sizes. This step resulted in 6,849 (Tobacco800), resp. 18,917 (Archive22k) features encoding raw frequency counts of all word types on each page. 

\paragraph{Topic composition:} In a second step, we obtained features of topical composition for each page from an unsupervised machine learning process. For this, we rely on Latent Dirichlet Allocation (LDA), also referred to as topic modeling \cite{Blei.2003}.\footnote{Actually, there is a large variety of unsupervised topic models as well as many other methods to reduce sparse, high-dimensional text data to a dense, lower-dimensional space (e.g. latent semantic analysis). For our baseline system, we stick to LDA as the seminal and most widely-used topic model.} Topic proportions based on multinomial posterior probability distributions $\theta$ from a topic model can be used as a dense feature vector comprising latent semantics of the modeled documents. In addition to  highly sparse n-gram features, they can provide useful information to any text classifier.
Following a method proposed by \cite{Phan.2011}, we presented single page texts as pseudo-documents to the process and compute a model with $K = 50$ (Tobacco800), resp. $K = 100$ (Archive22k) topics. Different topic resolutions were chosen again with respect to different collection sizes. For each page $p$, we then use the resulting topic-page distribution $\theta_p$ as feature vector supplementary to the previously extracted vector of unigram counts.

\paragraph{Topic difference:} We expect multi-page documents to comprise a rather coherent topic structure. Thus, for each page $p$, we determine the difference between its topic composition $\theta_p$ and its predecessor $\theta_{p-1}$ as a third feature type for PSS. We utilize two measures, Hellinger distance and Cosine distance, to create two additional features. While the former is a common metric to compare two probability distributions, the latter also has been adopted successfully to compare topic model results \cite{Niekler.2012a}. Distance values near zero indicate a high similarity of topic composition compared to the predecessor page. Values near one indicate a significant change of topic composition which could indicate the beginning of a new document. 

\paragraph{Predecessor pages:} As a last feature type, we add a copy of features extracted in the previous three steps belonging to the predecessor page as new features to each current page. This can be achieved easily by appending their values together with a new unique feature identifier. For this, we simply concatenate existing feature identifiers with a prefix, e.g. `PREV\#'. This is necessary to allow for the distinction between feature values for the current page and copied values from the predecessor page. By this, any classifier not only can rely on the information about characteristics of the current page for its decision but also may learn from information contained on the previous page. For instance, the presence of a salutation phrase such as ``With kind regards'' on a predecessor page highly increases the probability for the beginning of a new document on the current page.

The performance of \emph{SVM classification} to determine for each page whether it is the beginning of a new document or the continuation of the current document is tested in different steps. In each step, one of the four just introduced feature types is added to the feature set. Step-wise addition of the feature types to the linear SVM allows controlling whether each type effectively provides valuable information for the process.

\subsection{Neural networks on text and image features}
\label{subsec:cnn}

\begin{figure}
\centering
\caption{CNN + MLP architecture for PSS}
\label{fig:arch}
\includegraphics[width=0.5\textwidth]{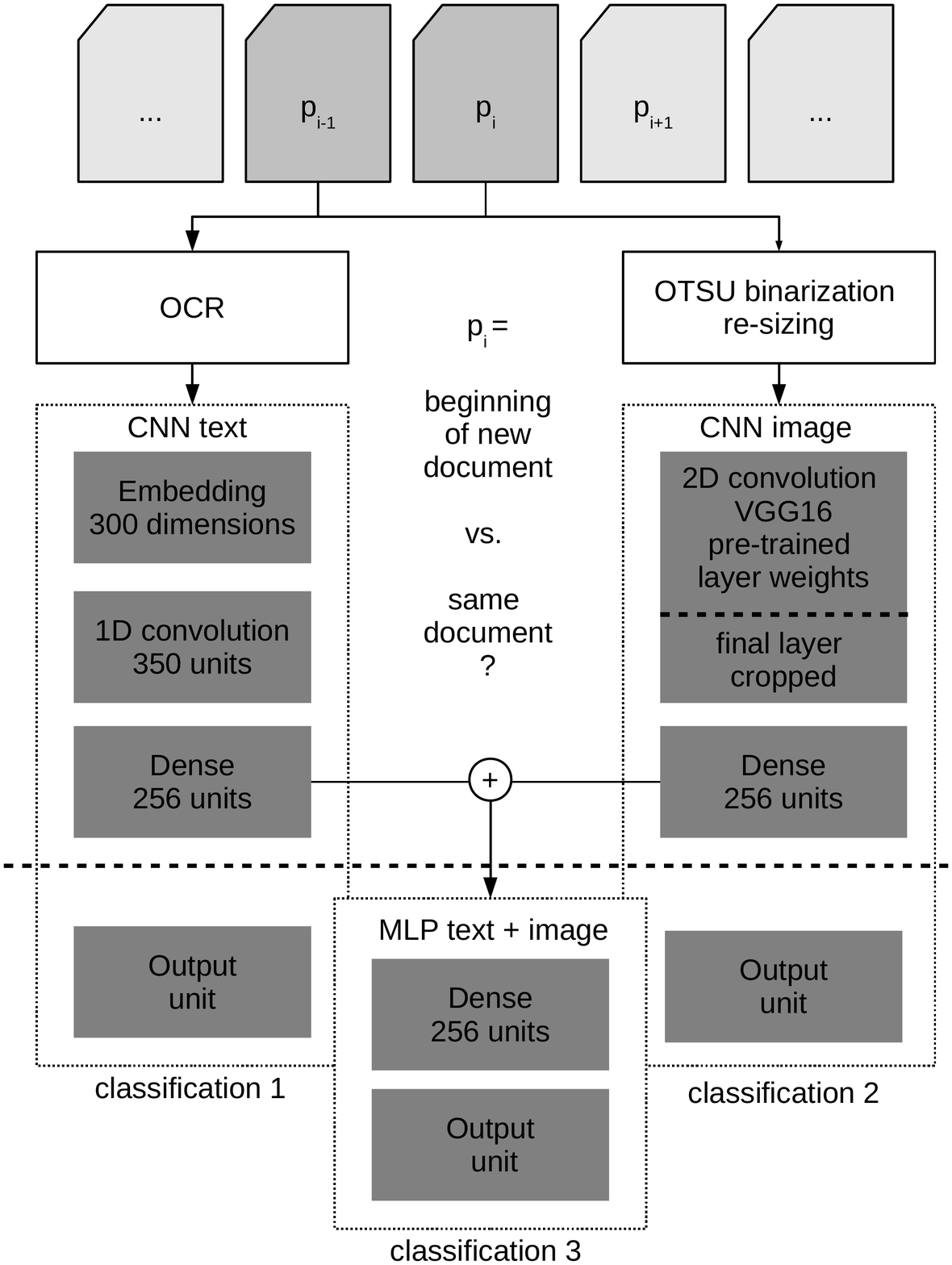}
\end{figure}

For our new PSS approach (cp. Fig. \ref{fig:arch} for a schematic representation of the architecture), we first create two separate convolutional neural networks (CNN) for binary classification of pages into either SD or ND, one based on text data and another based on image scans. In a third step, we combine the learned parameters from the two final hidden layers of both CNN to an input vector of features for a multi-layer perceptron. This MLP delivers a third and final classification result based on both feature types.

\paragraph{CNN for text data:}
\cite{Kim.2014} proposed a simple but effective CNN-architecture for text classification which achieved high performance for sentiment analysis tasks on standard data sets. He uses 1-dimensional convolution over word sequences encoded as embedding vectors.
We adopt a slightly simpler version of this network architecture by relying on only one kernel size instead of combining convolution layers with three different kernel sizes. Our network starts with an embedding layer with 300 dimensions, followed by a convolution layer with 350 filters and a kernel size of 3. On the resulting convolution filters, global max pooling is applied, followed by a dense layer of 256 neurons with ``ReLU'' activation, a dropout layer ($dr=.5$) and a final prediction layer for the binary class (sigmoid activation). The embedding layer is randomly initialized.\footnote{High performance for sentiment analysis in \cite{Kim.2014} is achieved by initializing the embedding layer with pre-trained word2vec embeddings \cite{Mikolov.2013} obtained from very large empirical data sets. Since we operate with data from two different languages, do not classify for semantic categories such as sentiments and also have a situation of rather noisy OCR data, we refrained from using pre-trained word embeddings in our setup.} Learning for this network was performed using RMSProp optimization with learning rate 0.0002 and mini-batches of size 32.

\paragraph{CNN for image data:}
Following the works in \cite{Noce.2016} and \cite{Gallo.2016}, we use a very deep CNN architecture to classify scanned pages based on their binarized and resized representation as 224$\times$224 pixels. We employ a network of 16 weight layers with very small convolution filters ($3 \times 3$) and max pooling as introduced by \cite{Simonyan.2014}. The network is initialized with pretrained weights based on the `imagenet' dataset (VGG-16). Actually, `imagenet' provides manually labeled photographs for object recognition tasks. But, earlier work has shown that CNN weights pre-trained on imagenet, although not specifically intended for the task of document image classification, can significantly improve DIC results for small datasets, too \cite{Harley.2015}. Hence, we expect them to be beneficial also for our PSS task.

To allow the network to adapt to our specific data and classification task, we applied a common technique of fine-tuning pretrained deep CNN. For this, we removed the final prediction layer and flattened the output of the last fully connected layer. Then, we fixate all weights of the original model layers. On top of this architecture, we added a new trainable, fully connected layer with 256 units and dropout regularization ($dr = .5$), and a new final prediction layer (sigmoid activation) for our binary classification task. Learning for this network was performed using the Adam optimizer with a small learning rate ($lr=0.0001$) and mini-batches of size 32.

\paragraph{Combining text and visual features:}

Each of the two previously introduced CNN are capable of classifying pages into either SD or ND on their own. But, since different information is utilized in each approach, we expect a performance gain from combining textual and visual information. For this, we modify the two previously introduced models in the same way. First, each model is trained on the training data individually. Then, we remove the final prediction layer from each model. In a next step, each example from the training and test data is fed into the networks again, to receive prediction values from the last fully connected layers of the two pruned networks. The output values from these last layers can be interpreted as new feature vectors for each data instance which encode dimensionality-reduced information from both, text and images.

From the text-based CNN, we receive a feature vector of 256 dimensions for each page according to the last dense layer of the model. To this vector, we concatenate the $K$ inferred topic proportion features and the two topic distance features from our baseline approach. Since text features from predecessor pages proved to be very useful in SVM baseline classification, we also use features from neighbor pages in our final model.
For this, we concatenate the vector of the current page with the vectors from its two predecessor pages to one text feature vector of length $3\times(256+K+2)$. In a last step, we concatenate the 256 image features from the image-based CNN to receive a final vector of 1,180 (Tobacco800), resp. 1,330 (Archive22k) dimensions.\footnote{Concatenating text features from a window size 3 has been decided experimentally. We also found that concatenating image features from predecessor pages did not improve the final performance.}

These final feature vectors now encode both, text and visual information from each page. They serve as input for a new MLP network consisting of 256 fully connected nodes with ``ReLU'' activation and l2-regularization ($factor = 0.01$), followed by dropout regularization ($dr = .5$) and a final, fully connected prediction layer  with sigmoid activation. Learning is performed using the Adam optimizer ($lr = 0.0005$) and a batch size of 16.

\section{Evaluation}
\label{sec:eval}

\textbf{Quantitative evaluation:} Table~\ref{tab:evaluation} displays the results of all tested model architectures and features types for PSS on our two investigated data sets.
Performance is measured by the accuracy of identification of a new document beginning vs. continuity of the same document. Since the distribution of both classes is fairly uneven due to different document length (there are a lot more pages in the SD class), we additionally use kappa statistics to report a chance-corrected agreement between human and machine separations of page streams.

\begin{table}
\centering
\caption{Evaluation of page stream segmentation}
\label{tab:evaluation}
\begin{tabular}{lrrrr}
\textbf{Approach/dataset} & \multicolumn{2}{c}{\textbf{Archive22k}} & \multicolumn{2}{c}{\textbf{Tobacco800}} \\
                          & Acc.             & kappa             & Acc.             & kappa            \\ \hline
SVM unigrams                  & 0.840                & 0.421             & 0.829                & 0.640            \\
+ topic composition                     & 0.839                & 0.419             & 0.829                & 0.640            \\
+ topic difference            & 0.847                & 0.446             & 0.837                & 0.657            \\
+ predecessor page        & 0.855                & 0.446             & 0.822                & 0.624            \\ \hline
CNN Text                  & 0.904                & 0.594             & 0.760                & 0.493            \\
CNN Image                 & 0.884                & 0.515             & 0.837                & 0.654            \\ \hline
MLP Image + Text          & \textbf{0.929}                & \textbf{0.691            } & \textbf{0.911}                & \textbf{0.816}           
\end{tabular}
\end{table}

The text features specifically engineered for PSS based on LDA topic composition and difference between consecutive pages improve the SVM results for text-based classification. Adding features from the predecessor page improves results for one dataset (Archive22k), but not for the other (Tobacco800). 

For the German dataset, we can observe that document boundaries can be identified more accurately with the CNN architectures than with linear SVM classification. For the English dataset, SVM constantly beats convolutional neural net classification on text features, but not on image features. One potential reason might be the rather small size of the dataset which does not contain enough examples for the complex CNN architecture to learn from. 

For both datasets, accuracy and kappa statistics improve significantly when image and text feature types are combined in one MLP architecture. The classifier achieves circa 93~\% accuracy on the German dataset and more than 91~\% on the English data. Compared to the results reported by the studies in section~\ref{sec:related_work}, this can be regarded as a new state-of-the-art for page stream segmentation.

\textbf{Qualitative evaluation:} Although first pages and subsequent pages of documents can be distinguished with high accuracy, our improved PSS approach still makes a considerable number of errors. There are two types of errors for the binary classification of pages: False positives (FP) and false negatives (FN). According to the manually separated pages in the gold standard, FP are subsequent pages (class SD) that are recognized by the classifier as first page (class ND). FN are defined the other way around. 
For SVM classification in the German dataset, FP account for about three quarters of all errors in the test set. FN make up about one quarter of all errors. This mismatch means that automatic PSS potentially splits the page stream into more documents than there are actually in the gold standard. 

For the final MLP architecture, we observe not only an increased accuracy but also a more balanced ratio between FP and FN. Apparently, the architecture is able to avoid more FP errors that FN errors resulting in less (incorrect) document splits.
On closer inspection, the remaining FPs often prove to contain characteristics of valid first pages, e.g. the beginning of a sub-document attached to one main document. This means although an automatic split is counted as an error in the quantitative evaluation, it nevertheless can represent a meaningful, content-related split for our application of retro-digitizing a large paper archive.

\section{Discussion}

We presented a new approach for page stream segmentation based on binary classification of pages. Our approach combines two convolutional neural networks to create features from image and text data which are used as input for a third MLP network. Our approach achieves very high accuracy for the task  to identify the beginning of a new document in a flow of scanned document pages. An accuracy above 91~\% for the Tobacco800 dataset which has been used in previous studies on this task, and accuracy of 93~\% on our own dataset can be regarded as a new state-of-the-art for this task.
The approach allowed us to drastically reduce costs for separating batch-scanned pages into document units in our project of retro-digitizing a research archive of around one million pages.

\section{Acknowledgements}

This work has been realized at the Leipzig University in the joint research project ``Knowledge Management of Legacy Documents in Science, Administration and Industry'' together with the Helmholtz Research Centre for Environmental Health in Munich and the CID GmbH, Freigericht. The authors thank colleagues at Helmholtz and CID for their valuable support.

\section{Bibliographical References}
\label{main:ref}

\bibliographystyle{lrec}
\bibliography{references}


\end{document}